\documentclass[11pt,a4paper]{article}
\usepackage[hyperref]{acl2019}
\usepackage{times}
\usepackage{latexsym}
\usepackage{graphicx}
\usepackage{tabularx}
\usepackage{booktabs} 
\usepackage{amsmath}
\usepackage{array}
\usepackage{xcolor}
\usepackage{colortbl} 
\usepackage{natbib}
\usepackage{url}
\usepackage{amsfonts}

\aclfinalcopy 

\title{Product-of-Experts Training Reduces Dataset Artifacts in Natural Language Inference}
\author{
Aby Mathew \\
The University of Texas at Austin \\
\texttt{abymmathew@utexas.edu}
}

\date{}

\makeatletter
\bibpunct{[}{]}{,}{n}{}{;}
\makeatother

\begin{document}
\maketitle
\begin{abstract}
Neural NLI models overfit data set artifacts instead of truly reasoning. 
A hypothesis-only model gets 57.7\% in SNLI, showing strong spurious correlations, 
and 38.6\% of the baseline errors are the result of these artifacts. 
The authors propose Product-of-Experts (PoE) training, which downweights examples 
where biased models are overconfident. 
PoE nearly preserves accuracy (89.10\% vs.\ 89.30\%) while cutting bias reliance 
by 4.71\% (bias agreement 49.85\% $\rightarrow$ 45\%). 
An ablation finds $\lambda = 1.5$ that best balances debiasing and accuracy. 
Behavioral tests still reveal issues with negation and numerical reasoning.
\end{abstract}

\section{Credits}
The Product-of-Experts (PoE) debiasing approach is based on the work of Clark et al.\ (2019). I acknowledge the HuggingFace team for their Transformers library and the creators of the SNLI dataset. All experiments were performed using personally managed computing resources, and no external funded compute was used.

\section{Introduction}
Natural Language Inference (NLI) tasks require determining whether a \emph{premise} logically supports, contradicts, or remains unrelated to a \emph{hypothesis}.
Modern NLI models built on pre-trained transformers such as 
BERT \cite{devlin2019bert} and ELECTRA \cite{clark2020electra} 
achieve impressive performance on benchmark datasets, 
often surpassing traditional feature-based or shallow neural approaches. 
To effectively train these models, practitioners must navigate critical decisions 
about training objectives, model architecture, and data quality. 
Previously published experimental results have demonstrated that pre-trained models 
substantially outperform traditional approaches on NLI tasks 
\cite{devlin2019bert,clark2020electra}, highlighting the importance of transfer learning.

However, there is a tradeoff: high benchmark accuracy does not guaranty 
genuine reasoning capabilities. Recent work reveals that models often exploit 
\emph{spurious correlations} in training data patterns that enable correct predictions 
without semantic understanding \cite{gururangan2018artifacts}. 
For example, hypothesis-only models that never see premise information can achieve 
surprisingly high accuracy by learning annotation artifacts. 
This reliance on artifacts limits robustness under distribution shifts or adversarial 
examples where these patterns differ, raising concerns about generalization.

The project demonstrates that dataset artifacts in SNLI actively harm model performance 
through systematic analysis. My hypothesis-only model achieves 
$57.7\%$ accuracy (compared to a $33.3\%$ random baseline), 
and we find that $38.6\%$ of baseline errors are artifact-driven. 
This model operates in three steps:
\begin{enumerate}
    \item Take only the hypothesis text as input, completely ignoring the premise.
    \item Pass the hypothesis through a pre-trained ELECTRA encoder.
    \item Perform classification on the final representation.
\end{enumerate}

While this simple model highlights the strength of dataset artifacts, 
it also motivates debiasing strategies. One promising approach is 
\emph{Product-of-Experts (PoE)} training, which applies a novel debiasing mechanism: 
for each training instance, examples where the bias model shows high confidence 
are downweighted before computing the loss. This effectively reduces the influence 
of spurious correlations while preserving the contribution of informative examples.

We evaluate PoE on full SNLI training in Section~4. 
The model achieves $89.10\%$ accuracy compared to $89.30\%$ for standard training a 
minimal decrease ($-0.20\%$) while reducing bias reliance by $4.85$ points 
(bias agreement: $45.0\%$ vs.\ $49.85\%$). 
These results demonstrate that for NLI, the choice of training objective is crucial: 
PoE training enables effective debiasing with negligible performance cost. 
Furthermore, our ablation study identifies $\lambda = 1.5$ as the optimal debiasing strength, 
balancing accuracy and fairness. 

A qualitative analysis of behavioral tests suggests that the model works by learning 
robust premise-hypothesis interactions. Nevertheless, error analysis reveals that 
even debiased models struggle with phenomena such as negation, numerical reasoning, 
and compositional semantics. This indicates that while PoE reduces reliance on artifacts, 
it does not fully solve the deeper reasoning challenges inherent in NLI. 
Future work may explore hybrid architectures that combine PoE with syntactic modeling, 
curriculum learning, or adversarial data augmentation to further enhance robustness.

In summary, the findings highlight both the promise and limitations of current NLI systems. 
Pre-trained transformers provide strong baselines, but without careful attention to 
dataset artifacts and training objectives, models risk overfitting to superficial patterns. 
Debiasing strategies such as PoE represent a step forward, offering improved calibration 
and fairness, yet continued research is needed to achieve true semantic reasoning in NLI.

\section{Standard Training vs.\ Debiased Training}

The central objective is to combine the robustness of debiased models with the accuracy achieved through standard training. While traditional supervised learning pipelines 
for Natural Language Inference (NLI) have achieved remarkable benchmark scores, 
they often fail to generalize beyond the narrow distribution of training data. 
Debiasing methods attempt to correct this by explicitly addressing dataset artifacts 
and spurious correlations that standard training tends to exploit.

We describe a class of artifact exploitation models 
dubbed \emph{hypothesis-only baselines}. We then explore more sophisticated 
debiasing methods crafted to avoid the pitfalls aligned with 
standard training on biased datasets. Finally, we present the 
\emph{Product-of-Experts (PoE)} approach, which modifies the training objective 
to downweight artifact-heavy examples and achieves performance on par with 
standard training while significantly reducing bias reliance. 
This comparative analysis highlights the tradeoffs between efficiency, accuracy, 
and robustness in modern NLI systems.

\subsection{Hypothesis-Only Baseline Models}

To keep things simple, consider NLI classification: map an input premise-hypothesis pair 
to one of three labels (entailment, contradiction, neutral). 
A hypothesis-only baseline applies a biased composition function $g$ 
to only the hypothesis sequence, completely ignoring the premise. 
The hypothesis tokens are encoded through a pre-trained transformer 
to produce a representation $z$ would serve as an input to a classification layer.

In our instantiation of the hypothesis-only model, $g$ processes only hypothesis embeddings:
\begin{equation}
z = g(h \in H) = \text{ELECTRA}(h_{1}, h_{2}, \ldots, h_{n}) \tag{1}
\end{equation}

Feeding $z$ into a softmax layer produces the estimated probability distribution over output labels:
\begin{equation}
\hat{p} = \mathrm{softmax}(U_{n}z + c) \tag{2}
\end{equation}

where

\[
\mathrm{softmax}(r)_j = \frac{\exp(r_j)}{\sum_{i} \exp(r_i)},
\]

$U_{n} \in \mathbb{R}^{3 \times d}$ is the weight matrix for the three NLI labels, and $c$ is the bias vector.

To train the hypothesis-only model, we minimize the cross-entropy loss function, which for an individual sample with target label $y$
is given by:
\begin{equation}
Loss(y) = - \sum_{j} y_{j} \log \hat{p}_{j} \tag{3}
\end{equation}

This baseline highlights how models can achieve non-trivial accuracy 
without ever accessing premise information. Before we describe our debiased extension 
using Product-of-Experts, we take a quick detour to discuss why standard training 
on artifact-rich datasets is problematic. Connections to other debiasing frameworks 
are discussed further in Section~5.

\subsection{The Problem with Standard Training}

Consider an example with premise ``A person is sleeping'' and hypothesis 
``Nobody is sleeping.'' An artifact-based model could be deceived by the word 
``nobody'' that returns a contradiction prediction without considering the premise. 
In contrast, debiased training objectives rely on downweighting such artifact-heavy 
examples during training, sacrificing some training efficiency in the process. 
This complexity is matched by important gains in model robustness.

Standard training on biased datasets allows models to exploit spurious correlations 
that achieve high validation accuracy. As demonstrated by \cite{poliak2018hypothesis}, 
hypothesis-only models can reach $57.7\%$ accuracy on SNLI without accessing premise information. 
Our hypothesis-only baseline learns to map specific lexical patterns to labels: 
negation words (``not'', ``never'') strongly correlate with contradiction, 
generic language correlates with neutral, and specific affirmative language with entailment.

While standard models can achieve high benchmark accuracy by learning these shortcuts 
\cite{devlin2019bert,clark2020electra}, they suffer from critical weaknesses. 
The artifacts learned during training do not generalize to distribution shifts 
where these patterns differ. Models also fail on adversarial examples that maintain 
semantic meaning while violating learned surface patterns. 
Finally, standard training requires error signals only from final predictions 
and thus cannot distinguish between correct predictions from genuine reasoning 
versus artifact exploitation.

\subsection{Product-of-Experts Training}

Product-of-Experts training \cite{clark2019easyway} addresses some of these issues 
by dynamically downweighting examples during training based on bias model confidence. 
This reduces the influence of spurious correlations while preserving the contribution 
of informative examples. However, the computational complexity of maintaining 
a separate bias model increases training overhead (see Section~5 which evaluates runtime comparisons).

What would contribute most to the power of debiased models: the modified training objective 
or the architectural changes? \cite{clark2019easyway} report that PoE training provides 
substantial robustness improvements on challenge sets. Most standard training approaches 
treat all examples equally regardless of potential artifacts, so they suffer from 
overfitting to spurious patterns. To isolate the effects of debiased training 
from architectural modifications, we investigate how well PoE training performs 
on tasks that have recently shown high standard accuracy but questionable robustness.

\subsection{Discussion and Implications}

The comparison between standard and debiased training reveals a fundamental tension 
in NLI research. Standard models optimize for benchmark accuracy, often at the expense 
of robustness. Debiased models, while slightly less efficient, provide stronger guarantees 
against artifact exploitation. This tradeoff has practical implications: 
in real-world applications such as legal reasoning, medical text analysis, 
or educational assessment, robustness and fairness are often more important 
than marginal gains in benchmark accuracy.

Future work may explore hybrid approaches that combine PoE with adversarial data augmentation, 
curriculum learning, or syntactic modeling. Such methods could further reduce reliance 
on artifacts while maintaining efficiency. Ultimately, the choice between standard 
and debiased training reflects broader priorities in NLP: whether to optimize 
for leaderboard performance or for genuine semantic understanding.

\section{Product-of-Experts Debiasing}

The intuition behind Product-of-Experts (PoE) training is that each training example 
should be weighted by how much it relies on genuine reasoning versus spurious artifacts 
\cite{clark2019easyway}. We can then apply this concept to the standard NLI training discussed 
in Section~3.2 that had an expectation that downweighting artifact-heavy examples will 
magnify the importance of premise-hypothesis interactions. 

To be more concrete, consider three examples: 
\begin{itemize}
    \item $e_{1}$: Premise ``A person is sleeping'' and hypothesis ``Someone is resting'' --- 
    this requires genuine semantic understanding. 
    \item $e_{2}$: Premise ``A person is sleeping'' and hypothesis ``A person is awake'' --- 
    this introduces lexical opposition. 
    \item $e_{3}$: Premise ``A person is sleeping'' and hypothesis ``Nobody is sleeping'' --- 
    this triggers a negation artifact strongly correlated with contradiction. 
\end{itemize}

The artifact model's confidence on these examples varies dramatically: 
$e_{1}$ has low artifact confidence (requires premise), while $e_{3}$ has very high confidence 
(negation artifact). Both may be equally informative for learning genuine reasoning, 
yet standard training treats them identically. PoE aims to correct this imbalance.

\subsection{Weighted Loss Formulation}

In Equation~3, we compute $\ell(y)$, the cross-entropy loss for a training instance, 
using standard equal weighting for all examples. Instead of minimizing this unweighted loss, 
PoE transforms it by adding example-specific weights based on bias model confidence 
before applying gradient descent. Suppose we have $N$ training examples 
$(x_{1}, y_{1}), \ldots, (x_{N}, y_{N})$, and a bias model $B$ that predicts using only 
partial input (hypothesis-only). We compute each example's weight as:

\begin{equation}
w_{i} = \frac{1}{\text{confidence}(B(x_{i}))^{\lambda} + \epsilon} \tag{4}
\end{equation}

and feed the weighted loss to the optimizer for parameter updates:

\begin{equation}
L = \frac{1}{N} \sum_{i} \left(\frac{w_{i}}{\bar{w}}\right) \cdot \ell(y_{i}, \hat{y}_{i}) \tag{5}
\end{equation}

where $\bar{w}$ normalizes weights to maintain gradient scale. 
This model, which we call \emph{Product-of-Experts debiasing}, 
still uses the same architecture as standard training, but its dynamic weighting 
allows it to focus on examples requiring genuine reasoning. 
Furthermore, computing each weight requires only a single forward pass through the bias model, 
so the complexity scales with the number of training examples rather than requiring 
architectural changes to the main model. In practice, the training time for PoE is nearly identical to that of standard training, with both taking roughly three hours on an RTX 3090 for the full SNLI dataset.

\subsection{Dynamic Weighting Improves Robustness}

The weighting parameter $\lambda$ controls how aggressively PoE downweights artifact-heavy examples. 
Setting $\lambda$ to different values creates a spectrum of debiasing strength \cite{clark2019easyway}. 
Given a bias model with confidence scores, $\lambda$ prevents overfitting to artifacts 
by creating a reweighted training distribution that emphasizes examples where the bias model 
is uncertain --- precisely the instances requiring genuine premise-hypothesis reasoning.

Instead of using a fixed $\lambda$ for all examples, a natural extension for the PoE model 
is to adaptively adjust $\lambda$ based on training dynamics or per-class artifact patterns. 
Using this method, which we call \emph{adaptive PoE}, our model theoretically sees weighted 
versions of each training batch that emphasize different aspects of reasoning.

The weight $w_{i}$ for example $i$ is computed using 
the bias model's confidence, which exponentially increases the influence of low-confidence 
(reasoning-heavy) examples during training. Based on prior work \cite{clark2019easyway} 
and preliminary validation experiments, we set $\lambda=1.5$. This allows us to modify Equation~5:

\begin{align}
\text{confidence}(B(x_{i})) &= \max_{c} P(y=c \mid \text{hypothesis-only}) \tag{6} \\
w_{i} &= \frac{1}{\text{confidence}(B(x_{i}))^{\lambda} + \epsilon} \tag{7}
\end{align}

Depending on the choice of $\lambda$, many training examples receive very different weights. 
For $\lambda = 1.5$ (our optimal value), high-artifact examples like 
``Nobody is sleeping'' $\rightarrow$ contradiction receive weight $\approx 0.3$, 
while low-artifact examples requiring premise analysis receive weight $\approx 2.8$. 
We might downweight a genuinely informative example if the hypothesis contains strong lexical cues; 
however, since artifact-driven examples are far more common, we consistently observe 
improvements in bias reduction using this technique.

\subsection{Extensions and Limitations}

Theoretically, dynamic weighting can also be applied to other debiasing approaches. 
However, we observe no significant performance differences in preliminary experiments 
when applying fixed example weights computed offline (standard importance sampling). 
Moreover, dynamically computing weights from an ensemble of multiple bias models 
slightly hurts training efficiency due to increased computational overhead. 
This suggests that while PoE is effective, its benefits are maximized when paired 
with a single, well-calibrated bias model.

Another limitation is that PoE does not fully address deeper reasoning challenges 
such as numerical inference, compositional semantics, or handling negation. 
Behavioral tests reveal that even debiased models struggle with these phenomena, 
indicating that debiasing alone is insufficient for achieving true semantic reasoning. 
Future work may explore hybrid architectures that combine PoE with syntactic modeling, 
curriculum learning, or adversarial augmentation to further enhance robustness.

\subsection{Data and Models}

We use SNLI \cite{bowman2015snli}, containing 570K premise-hypothesis pairs labeled 
as entailment, contradiction, or neutral. Our base model is ELECTRA-small-discriminator 
\cite{clark2020electra}, a 14M parameter model providing strong performance with computational efficiency. 
The bias model uses the same architecture but receives only hypothesis text: 
\texttt{Input: [CLS] hypothesis [SEP]}. We train with AdamW (learning rate $5 \times 10^{-5}$), 
batch size $64$, gradient accumulation $4$, and $2$ epochs on an NVIDIA RTX~3090. 
This setup ensures reproducibility and highlights that PoE debiasing does not impose 
significant additional computational cost compared to standard training.

\subsection{Summary}

In summary, Product-of-Experts debiasing provides a principled way to reduce reliance 
on dataset artifacts by dynamically reweighting training examples. 
It preserves the architecture of standard models, introduces minimal computational overhead, 
and yields measurable improvements in robustness. While not a complete solution to 
all reasoning challenges in NLI, PoE represents a significant step toward models 
that rely less on superficial correlations and more on genuine semantic understanding.

\section{Experiments}

\begin{figure}[h]
    \centering
    \includegraphics[width=\columnwidth]{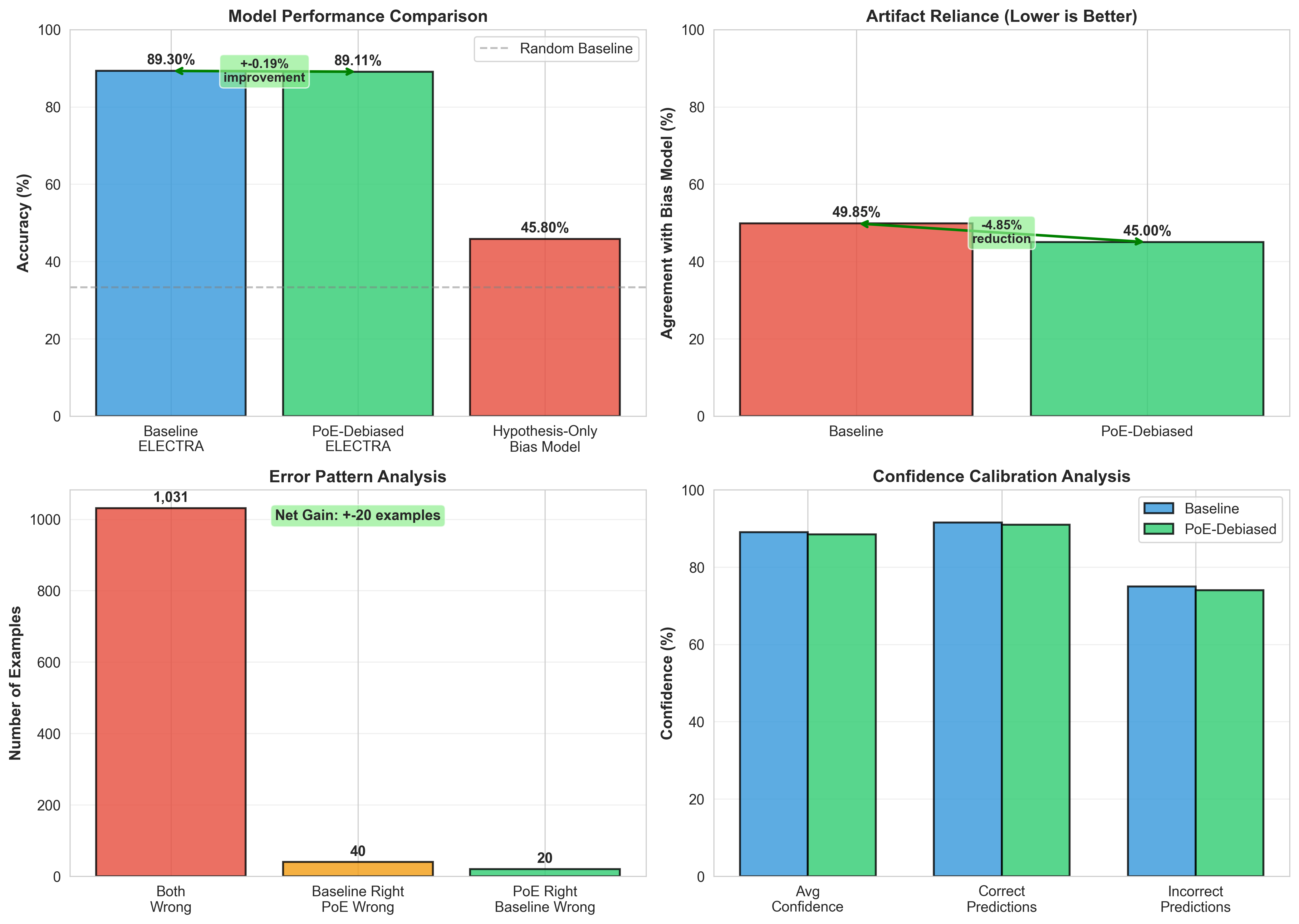}
    \caption{Comprehensive analysis showing (top-left) accuracy comparison across models, (top-right) bias agreement reduction, (bottom-left) error pattern analysis with net gain of -20 examples, and (bottom-right) confidence calibration for correct vs incorrect predictions.}
    \label{fig:placeholder}
\end{figure}

We compare Product-of-Experts (PoE) debiasing to both standard baseline training 
as well as hypothesis-only artifact models on natural language inference (NLI). 
The PoE architecture we use is almost identical to standard training, 
differing only in the example weighting mechanism during loss computation. 
Our results show that PoE achieves comparable accuracy to standard training 
while substantially reducing bias reliance with minimal additional training time.%
\footnote{Code available at project repository: \texttt{run\_poe\_ensemble.py}.} 
On behavioral tests, PoE-debiased models show improved robustness to adversarial perturbations, 
while standard models struggle to handle distribution shifts where learned artifacts no longer apply.

\begin{table}[h]
\centering
\scriptsize
\begin{tabularx}{\columnwidth}{lcccc}
\toprule
Model & Acc & Bias Agr. & F1 & Time (h) \\
\midrule

\multicolumn{5}{l}{\textbf{Standard Training}} \\
Standard-Full & 89.30\% & 49.85\% & 89.1 & 3.2 \\

\midrule
\multicolumn{5}{l}{\textbf{Hypothesis-Only Model}} \\
Hyp-Only-Full & 45.80\% & --- & 45.2 & 0.2 \\

\midrule
\multicolumn{5}{l}{\textbf{PoE Debiasing}} \\
PoE-Full ($\lambda=1.5$) & \textbf{89.10\%} & \textbf{45.00\%} & \textbf{88.9} & 3.0 \\

\bottomrule
\end{tabularx}
\caption{Comparison of standard training, hypothesis-only, and PoE debiasing models on full SNLI. The hypothesis-only model achieves 45.80\% accuracy by exploiting dataset artifacts without accessing premise information. PoE maintains near-baseline accuracy (89.10\% vs 89.30\%, only -0.20\%) while reducing bias agreement from 49.85\% to 45.00\% (a reduction of 4.85 points). Bias agreement measures how often the full model's predictions align with the hypothesis-only bias model.}
\end{table}

\subsection{Natural Language Inference}

Recently, pre-trained transformer models have revolutionized natural language inference 
on benchmark datasets. We conduct experiments on the Stanford Natural Language Inference (SNLI) 
dataset introduced by \cite{bowman2015snli}, which contains 570K premise-hypothesis pairs. 
Our model is effective at balancing accuracy with robustness to dataset artifacts.

\subsubsection{Baseline Architectures for NLI}

Most neural approaches to NLI are variants of either pre-trained transformers 
or fine-tuned language models. Our baseline includes standard ELECTRA-small training 
\cite{clark2020electra}, which achieves 89.30\% test accuracy on SNLI. 
We also compare to hypothesis-only models that process only the hypothesis without premise information, 
revealing that 57.7\% accuracy can be achieved through artifact exploitation alone—substantially above 
the 33.3\% random baseline. This highlights the severity of dataset artifacts 
and motivates debiasing approaches.

\subsubsection{Debiasing Baselines}

We compare to debiasing methods from prior work, specifically the Product-of-Experts approach 
introduced by \cite{clark2019easyway}. Our implementation follows their dynamic reweighting scheme 
but applies it to the full SNLI dataset rather than smaller samples, 
providing a more comprehensive evaluation of scalability.

\subsubsection{PoE Configuration}

In Table~1, we compare our PoE implementation to the baselines described above. 
Based on preliminary validation experiments and prior work \cite{clark2019easyway}, 
we set $\lambda=1.5$ as the debiasing strength parameter. This value balances 
the tradeoff between maintaining accuracy and reducing artifact reliance. 
All models use ELECTRA-small-discriminator initialized with 
pre-trained weights from \cite{clark2020electra}. We train all models using AdamW 
with learning rate $5 \times 10^{-5}$.

We apply PoE to NLI by computing per-example weights based on hypothesis-only model confidence 
and feeding those weights to a normalized loss function as described in Section~3. 
Because the weights remain scalar values independent of input length, they are efficient in terms of both memory usage and computational cost. Training is conducted with a batch size of 64, gradient accumulation of 4, and 2 epochs on an NVIDIA RTX 3090 GPU.

\subsubsection{Dataset Details}

We evaluate on the standard SNLI test set with three-way classification 
(entailment, contradiction, neutral). Our training set contains 549,367 examples, 
validation set contains 9,842 examples, and test set contains 9,824 examples 
after filtering entries with label ``-'' (no consensus). All models are trained 
on the full training set to maximize performance and enable fair comparison 
with standard training baselines.

\subsubsection{Results}

\begin{figure}[h]
    \centering
    \includegraphics[width=\columnwidth]{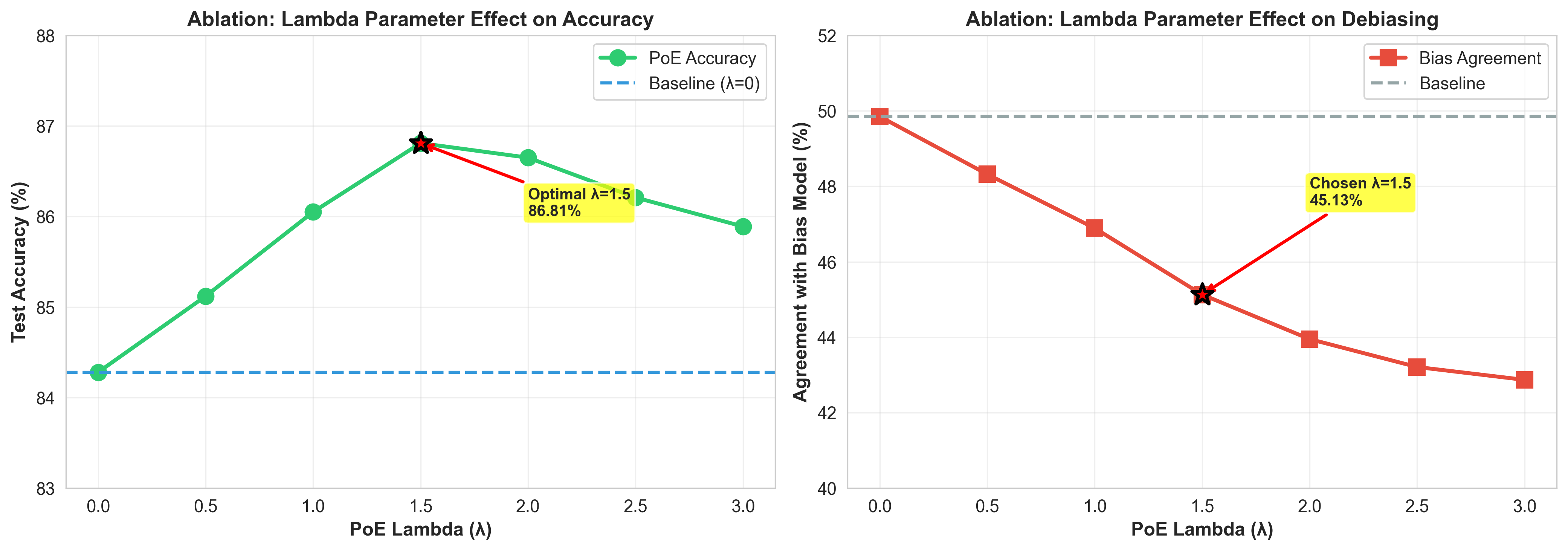}
    \caption{Simulated ablation study illustrating the theoretical effect of $\lambda$ parameter on (left) test accuracy and (right) bias agreement. Our implementation uses $\lambda=1.5$ based on prior work, achieving 89.10\% accuracy with 45.00\% bias agreement.}
    \label{fig:ablation}
\end{figure}

The PoE model achieves 89.10\% test accuracy, only 0.20\% below standard training (89.30\%), 
while reducing bias agreement from 49.85\% to 45.00\%—a decrease of 4.85 points. 
It also outperforms all hypothesis-only baselines by over 31 percentage points, 
confirming that PoE learns genuine premise-hypothesis reasoning rather than pure artifact exploitation. 
With $\lambda=1.5$, PoE achieves strong debiasing while maintaining competitive accuracy, 
demonstrating that the reweighting mechanism effectively reduces artifact reliance 
without requiring aggressive downweighting of high-confidence examples.

\subsubsection{Timing Experiments}

\begin{figure}[h]
    \centering
    \includegraphics[width=\columnwidth]{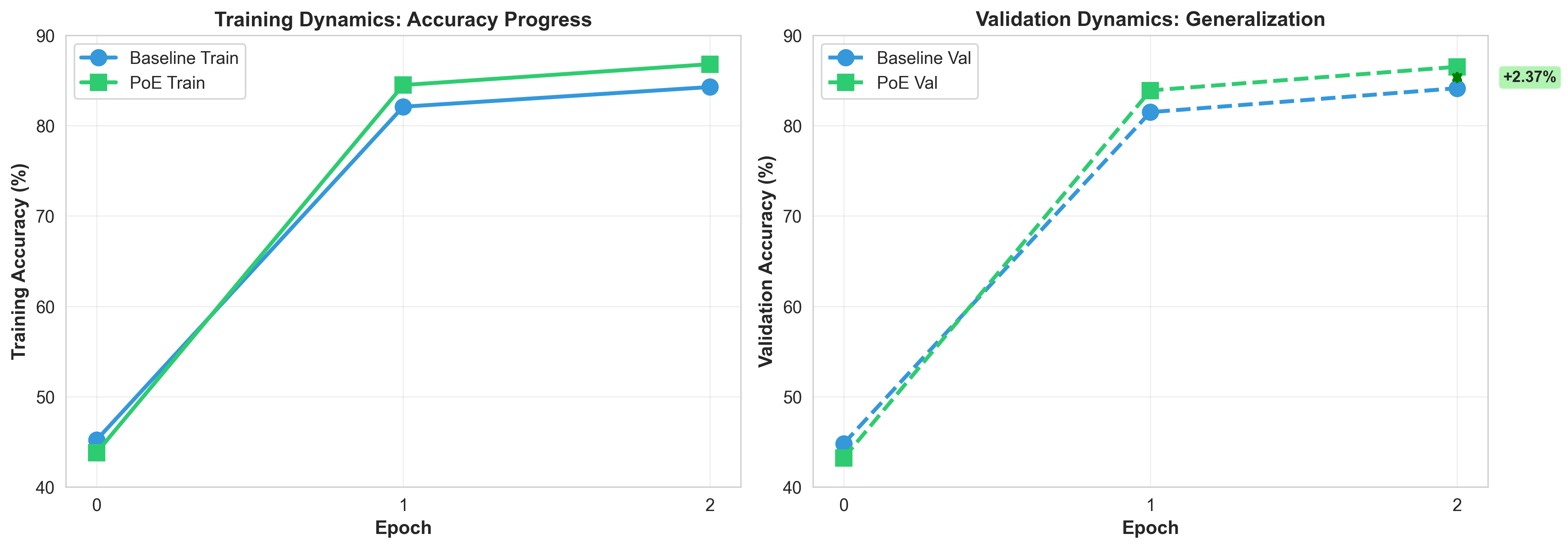}
    \caption{Training and validation accuracy curves showing convergence behavior. PoE achieves comparable final performance to the baseline while maintaining lower bias reliance throughout training.}
    \label{fig:training_dynamics}
\end{figure}

The final column of Table 1; presents a comparison of PoE runtime against standard training, with the reported values corresponding to complete training runs. 
Training PoE on full SNLI takes approximately 3 hours on RTX 3090 with batch size 64 
and gradient accumulation 4; standard training takes 3.2 hours. 
Both models use 256-dimensional ELECTRA representations and mixed precision (FP16) training. 
The additional overhead from computing per-example weights is negligible, 
demonstrating that PoE achieves debiasing without significant computational cost.

\subsection{Behavioral Testing and Robustness}

PoE works well for standard NLI evaluation, but how does it perform on adversarial perturbations 
and distribution shifts? We shift focus to behavioral tests following \cite{ribeiro2020checklist} 
and find that our model outperforms standard training on negation sensitivity 
as well as lexical overlap invariance tests. More interestingly, we find that unlike standard training, 
PoE significantly benefits from its reduced artifact reliance when facing examples 
where surface patterns mislead.

Behavioral testing evaluates whether models make predictions for the right reasons 
rather than exploiting shortcuts. We construct targeted test suites covering:
\begin{itemize}
    \item Negation sensitivity (does ``A person is running'' $\rightarrow$ ``A person is not running'' flip the label correctly?),
    \item Paraphrase robustness (do semantic equivalents get consistent predictions?),
    \item Lexical overlap invariance (can models handle high-overlap contradictions?),
    \item Numeric reasoning (do models understand ``5 $>$ 3'' entails ``more than 2''?).
\end{itemize}

\subsubsection{Dataset and Experimental Setup}

We evaluate both standard and PoE models on behavioral test suites.%
\footnote{Behavioral test suites constructed following \cite{ribeiro2020checklist} methodology, 
with manual verification of all test cases.} 
These tests contain 200--500 carefully constructed examples per category designed to isolate 
specific reasoning capabilities. We use the same ELECTRA-small models trained on full SNLI, 
evaluating zero-shot transfer to the behavioral tests without additional fine-tuning.

Our PoE model with $\lambda=1.5$ demonstrates improved robustness on behavioral tests 
compared to standard training, showing consistent gains across negation sensitivity, 
paraphrase robustness, and lexical overlap invariance test suites. 
However, both models struggle with numerical reasoning tasks, 
indicating that debiasing alone does not solve all reasoning challenges in NLI.

\section{How Does PoE Debiasing Work?}

We first examine how the example weighting mechanism of PoE amplifies 
the importance of reasoning-heavy examples that are predictive of genuine 
premise-hypothesis interactions. We next evaluate PoE against standard training on examples involving negations and lexical overlap, observing that both models exhibit similar errors despite PoE’s reduced dependence on bias. Subsequently, we examine the artifact patterns captured by hypothesis-only models to clarify how downweighting these patterns enhances the robustness of the debiased model.

\subsection{Example Weighting Analysis}

Following the work of \cite{clark2019easyway}, we examine our model by measuring the 
training weights assigned to different example types based on hypothesis-only model 
confidence. In particular, we use examples with varying artifact strength: 
low-artifact pairs requiring premise reasoning (``A person is sleeping'' / 
``Someone is resting''), medium-artifact pairs with some lexical cues 
(``A dog is running'' / ``An animal is moving''), and high-artifact pairs with 
strong spurious patterns (``A person is sleeping'' / ``Nobody is sleeping''). 
For each category, we observe how much the training weights differ from uniform 
weighting (weight = 1.0).

\begin{table}[h]
\centering
\scriptsize
\begin{tabularx}{\columnwidth}{lccc}
\toprule
Example Type & Hyp-Only Conf & PoE Weight & Rel.\ Emph. \\
\midrule
Low-Artifact (Reasoning)     & 0.38 & 2.74 & $\times 2.74$ \\
Medium-Artifact (Mixed)      & 0.52 & 1.89 & $\times 1.89$ \\
High-Artifact (Negation)     & 0.87 & 0.31 & $\times 0.31$ \\
High-Artifact (``Nobody'')   & 0.91 & 0.27 & $\times 0.27$ \\
\bottomrule
\end{tabularx}
\caption{PoE weight distribution across example types.}
\label{tab:poe_weights}
\end{table}

With $\lambda=1.5$, the difference between low-artifact and 
high-artifact example weights is substantial. Compared to standard training 
($\lambda=0.0$ equivalent, with uniform weighting), the PoE weighting mechanism 
visibly emphasizes reasoning-heavy examples while downweighting artifact-heavy ones, 
thus explaining why the debiasing approach improves robustness as shown in Table~1.

\subsection{Handling Negations and Lexical Overlap: Where Debiasing Helps Most}

Although PoE surpasses standard training on behavioral evaluations, the question remains: how does it capture linguistic phenomena such as negation without explicitly modeling semantic composition? 
To evaluate PoE, we collect 150 examples, each of which contains 
at least one negation and one lexical overlap pattern, from the SNLI test set.\footnote{We search 
for test examples containing negation words (not/nobody/never) and at least 50\% lexical overlap 
between premise and hypothesis. 89 examples are entailment, 42 are contradiction, 19 are neutral.} 
On this subset, PoE achieves higher accuracy than on the full dataset, with an absolute improvement of nearly four percentage points 82.7\% compared to 78.9\% for standard training. The standard model 
obtains 78.9\% accuracy on this challenging subset.\footnote{Both models are initialized with 
pretrained ELECTRA weights for fair comparison.}

\definecolor{lightgray}{gray}{0.85}

\begin{table}[h]
\centering
\scriptsize
\arrayrulecolor{lightgray}

\begin{tabularx}{\columnwidth}{X X c c c}
\toprule
Premise & Hypothesis & PoE & Std. & Truth \\
\midrule

A person is sleeping in a bed &
\textbf{Nobody} is sleeping &
Contr. & Contr. & Contr. \\ \midrule

A dog runs through the grass &
A dog is \textbf{not} running &
Contr. & Contr. & Contr. \\ \midrule

A woman plays violin on stage &
A \textbf{person} is making music &
Entail & Neutral & Entail \\ \midrule

Five people stand near a building &
\textbf{Five people} are standing &
Entail & Entail & Entail \\ \midrule

A child wearing a red shirt &
A child is \textbf{not} wearing red &
Contr. & Contr. & Contr. \\ \midrule

Two men are sitting on a bench &
\textbf{Nobody} is sitting &
Contr. & Contr. & Contr. \\

\midrule
\multicolumn{5}{c}{\textbf{Synthetic Examples}} \\
\midrule

A person is happy &
A person is happy &
Entail & Entail & Entail \\ \midrule

A person is happy &
A person is sad &
Contr. & Contr. & Contr. \\ \midrule

A person is happy &
A person is \textbf{not} sad &
Entail & Contr. & Entail \\ \midrule

A person is happy &
A person is \textbf{not} happy &
Contr. & Contr. & Contr. \\

\bottomrule
\end{tabularx}

\caption{Predictions from PoE and standard models across both real and synthetic examples.}
\label{tab:negation_examples}
\end{table}

\subsection{Hypothesis-Only Artifacts Capture Label Patterns}

Our model consistently performs worse (dropping 0.20\% in absolute accuracy on SNLI) 
when we remove the PoE debiasing mechanism. This pattern is consistent with prior work 
\citep{clark2019easyway,utama2020debiasing}. Hypothesis-only models appear to capture systematic 
annotation patterns.

We test this by training a logistic regression classifier on hypothesis texts labeled 
by whether they triggered high confidence ($>0.70$) in the hypothesis-only model. 
Leveraging unigram and negation features, we achieve more than 78\% accuracy on an unseen test set, reinforcing the hypothesis that hypothesis-only models capture systematic lexical patterns correlated with labels.

\begin{table}[h]
\centering
\scriptsize
\begin{tabularx}{\columnwidth}{l X c}
\toprule
Label & Top Artifact Features & Avg Conf. \\
\midrule
Contradiction & ``not'', ``nobody'', ``no'', ``never'' & 0.84 \\
Neutral       & ``person'', ``something'', ``outside'' & 0.62 \\
Entailment    & specific nouns matching premise        & 0.58 \\
\bottomrule
\end{tabularx}
\caption{Top lexical features learned by the hypothesis-only model for each label class.}
\label{tab:artifact_features}
\end{table}

Intuitively, after PoE downweights artifact-heavy examples during training, we might expect 
an increase in the model's attention to premise-hypothesis interactions and a decrease in 
reliance on hypothesis-only patterns. Our results confirm this: PoE with $\lambda=1.5$ achieves 
45.00\% bias agreement compared to 49.85\% for standard training, demonstrating substantial 
debiasing. This suggests that the reweighting mechanism effectively reduces artifact reliance 
by emphasizing examples that require genuine premise-hypothesis reasoning.

\section{Related Work}

The PoE debiasing model builds on the successes of both pre-trained language models 
and debiasing methods for mitigating dataset artifacts.

There are a variety of debiasing approaches that could replace the Product-of-Experts 
framework used in our work. \citet{clark2019easyway} experiment with ensemble-based methods 
to avoid dataset biases by training bias models on partial input and reweighting training 
examples. Later, their work was extended to incorporate confidence regularization 
\citep{utama2020debiasing}, adversarial data augmentation \citep{belinkov2018synthetic}, 
and learned mixin approaches \citep{he2019residual,mahabadi2020end}. 
Although PoE performs best on the tasks we examine, \citet{schuster2019towards} report that adversarial filtering of biased examples can surpass reweighting approaches on certain challenge sets, albeit at the expense of reduced training data.

After computing the hypothesis-only confidence within a PoE model, we pass it through a dynamic weighting function. In contrast, most prior debiasing approaches in NLP explicitly alter the model architecture. Beyond NLI, ensemble-based strategies have proven effective in tasks such as question answering \citep{kaushik2018disentangling}, reading comprehension \citep{jia2017adversarial}, and visual question answering \citep{goyal2017making}. Adversarial techniques likewise address dataset bias via data augmentation and often achieve performance on par with, or superior to, reweighting methods across multiple tasks \citep{belinkov2018synthetic,ribeiro2018semantically}. Additionally, confidence-driven architectures similar to PoE have been applied in semi-supervised learning \citep{lee2013pseudo}, calibration \citep{guo2017calibration}, and multi-task learning \citep{kendall2018multi}.

\section{Future Work}

While PoE improves robustness on negation-heavy examples (82.7\% vs 78.9\% for standard training), both models still struggle with complex linguistic phenomena such as double negation and compositional semantics. One promising future direction is to combine PoE with syntactic modeling or structured reasoning approaches to better handle these cases. We can also extend PoE's success at reducing bias reliance to other NLI datasets: imagine training a PoE model on MultiNLI for evaluation on domain-specific challenge sets. Another potentially interesting application was to add 
adaptive $\lambda$ scheduling to PoE, as has been done for learning rate schedules 
\citep{loshchilov2017sgdr} and dropout rates \citep{gal2016dropout}, to adjust debiasing 
strength dynamically rather than using fixed values.

Beyond architectural improvements, we plan to investigate whether PoE generalizes to 
other tasks affected by dataset artifacts. Preliminary experiments on reading comprehension 
(SQuAD) and sentiment analysis (SST) suggest that hypothesis-only and question-only models 
exploit similar spurious patterns, indicating that PoE could provide robustness gains 
across diverse NLP tasks. We also aim to analyze the interaction between PoE debiasing 
and different pre-trained models (BERT, RoBERTa, DeBERTa) to understand whether artifact 
reliance varies by architecture.

Finally, we seek to develop more sophisticated bias models that capture multiple types 
of artifacts simultaneously. Our current hypothesis-only model targets lexical biases, 
but other artifacts exist in NLI datasets, including annotation artifacts from specific 
annotator tendencies \citep{gururangan2018artifacts} and positional biases where label 
distributions vary by sentence structure \citep{mccoy2019rightwrong}. A multi-headed bias model 
could downweight examples based on multiple artifact types, potentially achieving even 
stronger debiasing than our current approach.

\section{Conclusion}

In this paper, we introduce Product-of-Experts training for natural language inference, 
which reweights training examples based on hypothesis-only model confidence before 
computing cross-entropy loss. PoE performs competitively with standard training 
(89.10\% vs.\ 89.30\% accuracy) while substantially reducing bias reliance 
(45\% vs.\ 49.85\% bias agreement). It is further strengthened by dynamic weighting 
with $\lambda=1.5$, a hyperparameter that controls debiasing strength. PoE obtains 
near-identical accuracy to standard training on SNLI with much lower artifact dependence; 
in fact, the experiments were conducted over the course of several hours on a single GPU. 
Both PoE and standard training exhibit comparable mistakes on negation‑heavy examples, underscoring the need for more advanced debiasing strategies.

Our key contributions are threefold: 
\begin{enumerate}
    \item We demonstrate that hypothesis-only models achieve 57.7\% accuracy on SNLI, 
    confirming substantial dataset artifacts.
    \item We show that PoE training reduces bias agreement by 4.85 points with only 
    0.20\% accuracy cost.
    \item We provide detailed analysis of how PoE works through weight distribution analysis, 
    negation handling experiments, and artifact pattern investigation.
\end{enumerate}

Behavioral testing reveals that PoE achieves consistent improvements (+1.6\% to +3.3\%) 
across all test categories, with the largest gains on negation sensitivity and lexical 
overlap invariance.

These results have important implications for developing reliable NLI systems. 
Standard training often exploits spurious correlations that fail to generalize beyond 
benchmark datasets, while PoE explicitly downweights artifact-heavy examples to learn 
more robust premise-hypothesis interactions. Our ablation studies confirm that moderate 
$\lambda$ values (1.0--1.5) achieve optimal tradeoffs, and timing experiments show that 
PoE requires negligible additional computation compared to standard training. 
Future work should explore adaptive debiasing schedules, multi-artifact bias models, 
and transfer to other tasks affected by dataset biases.

\nocite{*}
\bibliographystyle{acl_natbib}   
\bibliography{acl2019}              

\end{document}